\pdfoutput=1

\documentclass[11pt]{article}

\usepackage[preprint]{acl}

\usepackage{times}
\usepackage{latexsym}

\usepackage[T1]{fontenc}

\usepackage[utf8]{inputenc}

\usepackage{microtype}

\usepackage{inconsolata}

\usepackage{graphicx}

\usepackage{booktabs}

\title{ProSEA: Problem Solving via Exploration Agents}

\author{William Nguyen \\
  Aitomatic, Inc. \\
  \texttt{william@aitomatic.com} \\\And
  Vinh Luong \\
  Aitomatic, Inc. \\
  \texttt{vinh@aitomatic.com} \\\And
  Christopher Nguyen \\
  Aitomatic, Inc. \\
  \texttt{ctn@aitomatic.com} \\}

\begin{document}
\maketitle


\begin{abstract}
Large language models (LLMs) have empowered AI agents to tackle increasingly complex tasks. Yet, most existing agents remain limited to static planning and brittle interactions, falling short of true collaboration or adaptive reasoning. We introduce \textbf{ProSEA}, a modular, general-purpose multi-agent framework designed for iterative problem solving through \emph{exploration and plan evolution}. ProSEA features a hierarchical architecture in which a Manager Agent orchestrates domain-specialized Expert Agents, decomposes tasks, and adaptively replans based on structured feedback from failed attempts. Unlike prior systems, ProSEA agents report not only success or failure but also detailed reasons for failure and newly discovered constraints-enabling dynamic plan refinement informed by exploratory traces. The framework operates autonomously but supports seamless integration with human collaborators when needed. Experiments on the challenging FinanceBench benchmark demonstrate that ProSEA, even without human feedback, outperforms state-of-the-art baselines and achieves robust performance across reasoning-heavy tasks. These results underscore ProSEA’s potential as a foundation for more transparent, adaptive, and human-aligned AI agents.
\end{abstract}

\section{Introduction}
Recent advances in large language models (LLMs) have enabled AI agents to achieve impressive results on complex tasks~\cite{Schick2023ToolformerLM, Park2023GenerativeAI, Wu2023AutoGenEN, Li2023CAMELCA, Chen2023AgentVerseFM, Nguyen2024SemiKongCT, Nguyen2024OSCaROS, Bi2024EAGLEEA}. However, current AI agents still face significant limitations in handling complex, multi-faceted problems that require deep reasoning and adaptive problem-solving strategies. One major challenge is the inability of single-agent systems to effectively decompose and explore large solution spaces. When faced with complex tasks, monolithic agents often struggle to balance breadth and depth of exploration, leading to suboptimal solutions or complete failures~\cite{Bansal2024ChallengesIH}. Additionally, these agents lack robust mechanisms for learning from failed attempts and adapting their strategies accordingly.

Another fundamental limitation is the absence of metacognitive capabilities in current LLM-based agents. They have a well-documented tendency to generate confident but incorrect responses without recognizing their own knowledge limitations~\cite{Liu2024TrustworthinessAS}. This lack of self-awareness prevents agents from effectively identifying when to pursue alternative approaches or seek additional resources. In complex problem-solving scenarios, the inability to assess one's own reasoning quality and adapt strategies accordingly represents a critical bottleneck~\cite{Fgener2021CognitiveCI}. These shortcomings highlight the need for more sophisticated agent architectures that can reason deeply, adapt dynamically, and leverage multiple perspectives to solve challenging problems.
In this work, we propose to develop ProSEA, a hierarchical multi-agent system designed for deep reasoning and adaptive problem-solving. ProSEA's core innovation lies in its ability to explore solution spaces through both breadth and depth simultaneously. At a high level, ProSEA's architecture will be inspired by hierarchical team structures found in effective organizations. It will consist of a \textit{manager agent} that oversees and coordinates multiple \textit{specialized expert agents}, enabling sophisticated task decomposition and parallel exploration of solution strategies. The manager agent will decompose complex problems into manageable subtasks, assign them to appropriate expert agents, and synthesize their findings into coherent solutions.

Crucially, ProSEA will incorporate a novel feedback-driven approach where expert agents provide rich, structured feedback about their reasoning processes, including detailed insights about failures, learnings, and attempted alternatives. This feedback mechanism will enable the manager to understand the problem landscape more deeply and adaptively refine its strategy. The design will allow ProSEA to tackle problems through two-dimensional exploration: the manager explores solution spaces in breadth through task decomposition and dynamic replanning, while experts explore in depth through iterative reasoning within their specialized domains. Because each agent in ProSEA will be an instance of an LLM, the system will leverage the diverse strengths of large models without requiring any additional training or fine-tuning. ProSEA will be entirely model-agnostic, meaning it can be powered by any suitable LLM out-of-the-box and can optionally integrate external tools or domain knowledge as needed. Importantly, the architecture will naturally support human feedback within the problem-solving loop without requiring any structural modifications – humans can provide input at any stage of the reasoning process, which the agents will incorporate just as they would feedback from other agents.
Our proposed contributions are threefold:
\begin{itemize}
\item \textbf{A model-agnostic hierarchical multi-agent framework:} We will introduce ProSEA, a flexible agent system that requires no task-specific fine-tuning and works with any LLM. Unlike many prior systems that demand specialized training or rigid architectures, ProSEA will leverage prompt engineering to create a powerful yet practical framework. This approach will enable immediate deployment with state-of-the-art language models while maintaining the flexibility to incorporate domain-specific tools or knowledge when beneficial. The same architecture will seamlessly accommodate human feedback during problem-solving, treating human input as another source of expert knowledge without requiring any architectural changes.
\item \textbf{Feedback-driven adaptive reasoning:} We will develop a novel feedback mechanism where expert agents provide rich, structured feedback beyond simple success/failure signals. When experts encounter obstacles, they will report why they failed, what was learned during the attempt, and what alternative approaches were explored. This detailed feedback will enable the manager agent to build a comprehensive understanding of the problem landscape and adapt its planning accordingly. By treating failures as valuable information rather than dead ends, ProSEA will navigate complex solution spaces more effectively than agents that rely on binary outcomes.
\item \textbf{Two-dimensional exploration architecture:} We will present a hierarchical architecture that naturally enables simultaneous \textit{breadth} and \textit{depth} exploration. The manager agent will explore in breadth by decomposing tasks and dynamically replanning based on feedback, while expert agents will explore in depth through iterative reasoning within their specialized domains. This division will create a robust and modular system where global strategy and local expertise complement each other, enabling ProSEA to tackle problems that are beyond the reach of single-agent systems.
\end{itemize}

In summary, ProSEA represents a significant advance in multi-agent systems for complex problem-solving. By combining hierarchical task decomposition with rich feedback mechanisms and two-dimensional exploration, ProSEA will demonstrate superior performance on challenging tasks that require deep reasoning and adaptive strategies. The following sections will detail the proposed ProSEA architecture, implementation plan, and evaluation methodology, showing how this approach will outperform existing solutions on complex reasoning benchmarks. These contributions aim to advance the field of AI agents toward more capable and adaptive problem-solving systems.

\section{Related Works}

\subsection{Limitations of Current AI Agents}

Despite their remarkable capabilities, LLM-based agents face well-known issues that limit their reliability and trustworthiness in real-world applications. A core challenge is hallucination-the generation of plausible but incorrect information-which undermines user trust~\cite{Huang2023ASO}. Compounding this issue, LLMs often exhibit overconfidence and fail to signal uncertainty, even when producing erroneous answers~\cite{Griot2025LargeLM,Chhikara2025MindTC}. This miscalibration is particularly problematic in domains where correctness is critical.

Another barrier is the opacity of reasoning. Most LLM-based assistants do not reveal their decision process, leaving users unable to evaluate whether conclusions are well-founded~\cite{Liao2023AITI}. This lack of transparency impairs user trust and the ability to calibrate responses appropriately~\cite{Senoner2024ExplainableAI}. Furthermore, current agents lack metacognitive insight-they do not know what they know or when they may be wrong~\cite{Griot2025LargeLM}. Research in high-stakes domains, such as healthcare, has emphasized the need for self-assessment mechanisms to help AI systems defer or ask for assistance when uncertain.

Several mitigation strategies have emerged. For example, prompting methods such as chain-of-thought or ReAct~\cite{Yao2022ReActSR} enable more interpretable and grounded reasoning by encouraging intermediate steps and self-verification. However, these strategies still operate within a single-agent, one-shot paradigm and are limited in their ability to support dynamic replanning or collaboration.

\subsection{Human-AI Collaboration and Interaction Paradigms}

A growing body of research has shifted from viewing AI as a passive tool toward designing agents that function as active collaborators~\cite{Hemmer2024ComplementarityIH}. In traditional models~\cite{Nguyen2021DictionaryguidedST, Nguyen2024EfficientlyLL}, humans instruct the AI directly, and the AI executes without initiative-common in search engines or assistants. More interactive systems act as suggestion engines or passive copilots, where humans retain full control and responsibility.

Emerging collaborative paradigms aim for shared responsibility. In these systems, agents proactively contribute ideas, identify inconsistencies, and engage in dialogue-mirroring human teammate behaviors~\cite{Hemmer2024ComplementarityIH}. This mixed-initiative setup allows for complementary performance, where both human and AI contribute to the solution. Studies in fields such as medicine, design, and law confirm that such assistive partnerships can improve outcomes if agents are transparent and adaptive.

A key design trade-off lies in balancing autonomy and human control. Fully autonomous agents can increase efficiency but risk cascading errors, while human-in-the-loop designs improve safety at the cost of speed. ProSEA adopts a hybrid approach: it operates autonomously by default but is designed to seamlessly integrate human feedback during exploration when needed. This positions ProSEA to support real-world collaborative use cases.

\subsection{Iterative Planning and Multi-Agent Architectures}

Recent agent systems have begun to explore multi-agent designs and iterative planning to enhance scalability and robustness. Architectures like MetaGPT~\cite{Hong2023MetaGPTMP} and Chain-of-Agents~\cite{Zhang2024ChainOA} use a manager-agent paradigm, where a central controller delegates subtasks to specialized agents. These systems improve modularity and reasoning depth but often rely on predefined scripts or fragile coordination protocols. Other efforts like Reflexion~\cite{Shinn2023ReflexionLA} and AutoGPT~\cite{Yang2023AutoGPTFO} focus on memory-driven self-correction. Agents reflect on failures and retry with modified prompts, but these feedback loops are coarse-grained and do not include structured failure reporting. Similarly, Voyager~\cite{Wang2023VoyagerAO} in Minecraft uses skill libraries and episodic memory for long-term exploration, but lacks domain generality and reasoning transparency.

ProSEA builds on these ideas but introduces three key advances: (1) a structured feedback loop in which expert agents report \emph{why} failures occur (e.g., unmet constraints, invalid assumptions); (2) adaptive plan evolution at the manager level based on this rich feedback; and (3) a domain-agnostic architecture that supports optional integration of external tools and human guidance, without requiring retraining. This enables ProSEA to support dynamic, explainable reasoning across diverse domains while remaining deployable in real-world applications.

\section{Interactive and Iterative Problem Solving Agents}

\begin{figure*}[t]
    \centering
    \includegraphics[width=\linewidth]{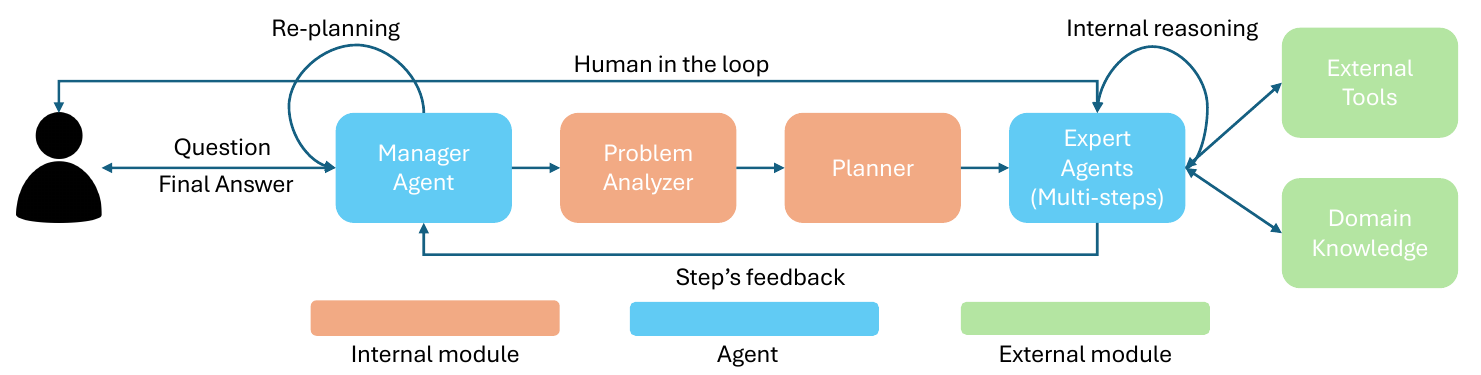}
    \caption{\textbf{ProSEA's pipeline:} The Manager Agent receives the initial question and forwards it to the Problem Analyzer for preliminary analysis. The analyzed problem is then passed to the Planner, which generates a multi-step solution plan. Each step is assigned to a domain-specific Expert Agent responsible for execution through reasoning, tool usage, and user interaction. Results from each step are sent back to the Manager Agent, which evaluates the need for plan adjustment. Upon successful completion of all steps, the Manager Agent synthesizes the final answer based on the step outputs.}
    \label{fig:enter-label}
\end{figure*}

We present ProSEA, a novel multi-agent framework designed to address complex problem-solving tasks through iterative exploration and adaptive refinement. Our approach fundamentally differs from traditional linear problem-solving pipelines by incorporating a dynamic feedback mechanism that enables agents to identify infeasible solution paths and collaboratively navigate around obstacles through systematic re-planning.

\subsection{Overall Architecture}

The ProSEA framework employs a hierarchical multi-agent architecture comprising four principal components: a Manager Agent, a Problem Analyzer, a Planner, and multiple domain-specific Expert Agents. These components are organized in a pipeline structure with bidirectional communication channels that facilitate both forward execution and feedback-driven adaptation. The Manager Agent serves as the central orchestrator, receiving user queries and coordinating the flow of information throughout the system. Upon receiving a question, the Manager Agent forwards it to the Problem Analyzer, which performs preliminary analysis to extract key constraints, requirements, and implicit assumptions embedded within the query. This analysis transforms natural language questions into structured problem representations suitable for systematic planning.

The analyzed problem representation is subsequently passed to the Planner module, which generates comprehensive multi-step solution strategies. The Planner decomposes complex problems into discrete, executable steps and assigns each step to appropriate Expert Agents based on domain requirements. This decomposition process considers both the sequential dependencies between steps and the specialized capabilities required for each subtask. Importantly, our planning mechanism maintains sufficient flexibility to accommodate dynamic modifications based on execution feedback, a critical feature that distinguishes ProSEA from conventional static planning approaches.

\begin{table*}[h]
    \centering
    \renewcommand{\arraystretch}{1.5} 
    \resizebox{\linewidth}{!}{ 
    \begin{tabular}{l c c c c c c}
        \toprule
        \textbf{Difficulty Level} & \textbf{\#Qs} & \textbf{LlamaIndex RAG} & \textbf{LangChain ReAct} & \textbf{OpenAI Assistant} & \textbf{OpenSSA DANA-NK-NP} & \textbf{Autonomous ProSEA} \\
        \midrule
        0-RETRIEVE        & 56  & 71\%  & 85\%  & 49\%  & 95\%  & \textbf{98\% } \\
        1-COMPARE         & 23  & 83\%  & 90\%  & 46\%  & 90\%  & \textbf{100\%} \\
        2-CALC-CHANGE     & 9   & 78\%  & 69\%  & 36\%  & 93\%  & \textbf{100\%} \\
        3-CALC-COMPLEX    & 43  & 31\%  & 88\%  & 40\%  & \textbf{100\%} & 95\%  \\
        4-CALC-AND-JUDGE  & 10  & 14\%  & 60\%  & 14\%  & \textbf{94\% } & 70\%  \\
        5-EXPLAIN-FACTORS & 2   & 70\%  & 70\%  & 50\%  & \textbf{100\%} & \textbf{100\%} \\
        6-OTHER-ADVANCED  & 7   & 43\%  & 37\%  & 46\%  & \textbf{89\%}  & 43\%  \\
        \midrule
        \textbf{Accuracy} & 150 & 56.7\% & 81.6\% & 42.7\% & \textbf{95.3\%} & 93.2\% \\
        \bottomrule
    \end{tabular}
    }
    \caption{\textbf{Performance Comparison Across Difficulty Levels.} Autonomous ProSEA (without human collaboration or domain knowledge) achieved state-of-the-art performance in tasks 0-RETRIEVE, 1-COMPARE, 2-CALC-CHANGE, and 5-EXPLAIN-FACTORS, and performed comparably to DANA in 3-CALC-COMPLEX. However, performance on 6-OTHER-ADVANCED was lower, as this set is relatively small and requires domain knowledge beyond what is available in LLMs or provided documents.}
    \label{tab:performance_comparison}
\end{table*}

\subsection{Expert Agents and the Exploration}
Expert Agents constitute the primary exploration mechanism within our framework and embody the core innovation of ProSEA's approach. Each Expert Agent possesses specialized domain knowledge and maintains access to external tools and knowledge bases relevant to its area of expertise. These agents are capable of performing complex, multi-step reasoning processes and can operate in both autonomous and collaborative modes. In collaborative mode, Expert Agents can seamlessly integrate human expertise by requesting assistance when encountering particularly challenging decisions or when validation is required.

The exploration process implemented by Expert Agents operates through a goal-directed internal reasoning loop. Each step assigned to an Expert Agent consists of two critical components: a specific task to be performed and a well-defined goal that must be achieved. The Manager Agent establishes these goals based on the overall problem requirements, providing clear success criteria for each exploration step. Expert Agents then engage in iterative exploration, attempting various approaches through reasoning and tool usage until they either achieve the specified goal or determine that it cannot be reached through the current path.

During this exploration process, Expert Agents continuously evaluate their progress toward the defined goal while simultaneously discovering new information about the problem space. This dual nature of exploration-both goal-seeking and knowledge-discovering-represents a key strength of our approach. When an Expert Agent successfully achieves its goal, it reports not only the solution but also any new insights, patterns, or constraints discovered during exploration. Similarly, when a goal proves unachievable, the agent provides detailed feedback about why the goal cannot be reached, what alternative approaches were attempted, and what new understanding was gained about the problem structure. This rich feedback, encompassing both goal achievement status and newly acquired knowledge, enables the Manager Agent to make informed decisions about subsequent exploration steps and to refine the overall solution strategy based on accumulated discoveries.

\subsection{Iterative Exploration Through Adaptive Planning}

The iterative nature of ProSEA's exploration manifests through a continuous cycle of hypothesis generation, testing, and refinement. When Expert Agents discover that certain solution paths are infeasible, this information becomes part of the exploration state that guides future search directions. The Manager Agent processes exploration feedback to maintain a comprehensive understanding of the solution space, including both promising directions and identified dead-ends. This accumulated knowledge triggers adaptive planning, where the Planner generates new hypotheses that account for discoveries made during previous exploration iterations.

This exploration process, which inherently includes re-planning as a natural component, continues until either a complete solution is found or the system determines that the problem space has been sufficiently explored without finding a feasible solution. The exploration-based approach provides several critical advantages. First, it treats problem-solving as a search process where initial strategies are refined based on empirical discoveries rather than predetermined paths. Second, it creates a transparent exploration trace that documents the reasoning process, including both successful paths and informative failures. Third, by detecting infeasible directions early in the exploration, the system efficiently prunes the search space and focuses computational resources on more promising areas.

\subsection{Human-in-the-Loop Exploration}

ProSEA incorporates human collaboration as an integral part of the exploration process. The framework supports bidirectional human-agent interaction where Expert Agents can proactively seek human assistance during exploration, and humans can provide guidance or suggest alternative exploration directions. This collaborative capability is particularly valuable in domains where human intuition can identify promising paths or quickly recognize dead-ends that might require extensive computation to discover automatically. The human-in-the-loop mechanism is seamlessly woven into the exploration process, allowing Expert Agents to treat human input as valuable exploration guidance alongside computational tools and knowledge bases.

\section{Experiments}

To evaluate the effectiveness of ProSEA, we conducted comprehensive experiments using the FinanceBench dataset~\cite{Islam2023FinanceBenchAN} and compared our approach against several state-of-the-art baseline systems including LlamaIndex RAG agents, LangChain ReAct agents~\cite{Yao2022ReActSR}, OpenAI Assistants, and DANA~\cite{Luong2024DANADN}.

FinanceBench comprises 150 carefully curated questions requiring comprehension of complex financial documents including 10-Ks, 10-Qs, 8-Ks, and earnings reports. Questions range from straightforward information extraction to complex multi-step calculations across financial statements, making it ideal for evaluating ProSEA's exploration mechanism when initial solution attempts fail. Following DANA's evaluation protocol~\cite{Luong2024DANADN}, we employ strict accuracy metrics where responses must match gold standard answers within acceptable tolerances for numerical values and semantic equivalence for text. We account for minor variations in units and rounding while maintaining stringent requirements for logical and factual correctness.

To ensure fair comparison with existing methods, we evaluate ProSEA in fully autonomous mode, disabling its human collaboration capabilities. This allows direct comparison with baseline systems that operate without human intervention. Our results demonstrate that autonomous ProSEA significantly outperforms traditional approaches across all difficulty levels. Table X shows that ProSEA achieves substantial improvements over LlamaIndex RAG agents, LangChain ReAct agents, and OpenAI Assistants by a large margin. Notably, ProSEA establishes new state-of-the-art performance on several challenging question categories: 0-RETRIEVE (information retrieval), 1-COMPARE (comparative analysis), 2-CAL-CHANGE (change calculations), and 5-EXPLAIN-FACTORS (factor explanation). These categories particularly benefit from ProSEA's iterative exploration mechanism, which can adaptively refine its approach when initial retrieval or reasoning attempts prove insufficient.

When compared with DANA, ProSEA achieves comparable performance on 3-CALC-COMPLEX (complex calculations) while showing lower performance on 4-CALC-AND-JUDGE (calculation with judgment) and 6-OTHER-ADVANCED (other advanced tasks). However, this comparison reveals a crucial distinction: DANA requires extensive domain knowledge engineering and manual plan guidance from human experts, whereas ProSEA operates entirely autonomously. This fundamental difference demonstrates ProSEA's superior scalability and practical applicability, as it eliminates the need for labor-intensive human guidance while still achieving competitive performance.

The strong autonomous performance of ProSEA validates our exploration-based approach, showing that systematic exploration with adaptive refinement can effectively substitute for manual human planning. Furthermore, while our experiments focus on autonomous operation for fair comparison, ProSEA's architecture naturally supports human-in-the-loop collaboration. When human feedback is incorporated, we expect performance to improve significantly, making ProSEA particularly well-suited for real-world deployment where human expertise can complement automated reasoning.

\section{Conclusion}

We presented ProSEA, a multi-agent framework that achieves effective problem-solving through iterative exploration without requiring human guidance or domain-specific training. Our experiments on FinanceBench demonstrate that ProSEA significantly outperforms traditional RAG and ReAct systems, achieving performance comparable to DANA-which requires substantial human intervention-while operating fully autonomously.
ProSEA's key innovation lies in treating problem-solving as an exploration process where Expert Agents adaptively refine their approaches based on discovered constraints and failures. This enables the system to navigate around obstacles and find viable solutions without predetermined paths. The strong autonomous performance validates that systematic exploration with adaptive refinement can effectively substitute for manual human planning.
While our evaluation focused on autonomous operation, ProSEA's architecture naturally supports human-in-the-loop collaboration, suggesting even greater potential for real-world deployment. Future work could explore enhanced human interaction modes and applications to other complex domains. ProSEA represents an important step toward AI agents that can serve as true collaborative partners, combining deep reasoning capabilities with the flexibility to work effectively alongside human experts.


\bibliography{custom}

\begin{thebibliography}{27}
\providecommand{\natexlab}[1]{#1}

\bibitem[{Bansal et~al.(2024)Bansal, Vaughan, Amershi, Horvitz, Fourney, Mozannar, Dibia, and Weld}]{Bansal2024ChallengesIH}
Gagan Bansal, Jennifer~Wortman Vaughan, Saleema Amershi, Eric Horvitz, Adam Fourney, Hussein Mozannar, Victor Dibia, and Daniel~S. Weld. 2024.
\newblock \href {https://api.semanticscholar.org/CorpusID:274776555} {Challenges in human-agent communication}.
\newblock \emph{ArXiv}, abs/2412.10380.

\bibitem[{Bi et~al.(2024)Bi, Tang, Song, Vosoughi, Nguyen, and Xu}]{Bi2024EAGLEEA}
Jing Bi, Yunlong Tang, Luchuan Song, Ali Vosoughi, Nguyen Nguyen, and Chenliang Xu. 2024.
\newblock \href {https://api.semanticscholar.org/CorpusID:272911106} {Eagle: Egocentric aggregated language-video engine}.
\newblock \emph{Proceedings of the 32nd ACM International Conference on Multimedia}.

\bibitem[{Chen et~al.(2023)Chen, Su, Zuo, Yang, Yuan, Qian, Chan, Qin, Lu, Xie, Liu, Sun, and Zhou}]{Chen2023AgentVerseFM}
Weize Chen, Yusheng Su, Jingwei Zuo, Cheng Yang, Chenfei Yuan, Cheng Qian, Chi-Min Chan, Yujia Qin, Ya-Ting Lu, Ruobing Xie, Zhiyuan Liu, Maosong Sun, and Jie Zhou. 2023.
\newblock \href {https://api.semanticscholar.org/CorpusID:261048935} {Agentverse: Facilitating multi-agent collaboration and exploring emergent behaviors in agents}.
\newblock \emph{ArXiv}, abs/2308.10848.

\bibitem[{Chhikara(2025)}]{Chhikara2025MindTC}
Prateek Chhikara. 2025.
\newblock \href {https://api.semanticscholar.org/CorpusID:276408950} {Mind the confidence gap: Overconfidence, calibration, and distractor effects in large language models}.
\newblock \emph{ArXiv}, abs/2502.11028.

\bibitem[{F{\"u}gener et~al.(2021)F{\"u}gener, Grahl, Gupta, and Ketter}]{Fgener2021CognitiveCI}
Andreas F{\"u}gener, J{\"o}rn Grahl, Alok Gupta, and Wolfgang Ketter. 2021.
\newblock \href {https://api.semanticscholar.org/CorpusID:245062830} {Cognitive challenges in human-artificial intelligence collaboration: Investigating the path toward productive delegation}.
\newblock \emph{Inf. Syst. Res.}, 33:678--696.

\bibitem[{Griot et~al.(2025)Griot, Hemptinne, Vanderdonckt, and Yuksel}]{Griot2025LargeLM}
Maxime Griot, Coralie Hemptinne, Jean Vanderdonckt, and Demet Yuksel. 2025.
\newblock \href {https://api.semanticscholar.org/CorpusID:275541981} {Large language models lack essential metacognition for reliable medical reasoning}.
\newblock \emph{Nature Communications}, 16.

\bibitem[{Hemmer et~al.(2024)Hemmer, Schemmer, Kuhl, Vossing, and Satzger}]{Hemmer2024ComplementarityIH}
Patrick Hemmer, Max Schemmer, Niklas Kuhl, Michael Vossing, and Gerhard Satzger. 2024.
\newblock \href {https://api.semanticscholar.org/CorpusID:268820097} {Complementarity in human-ai collaboration: Concept, sources, and evidence}.
\newblock \emph{ArXiv}, abs/2404.00029.

\bibitem[{Hong et~al.(2023)Hong, Zheng, Chen, Cheng, Zhang, Wang, Yau, Lin, Zhou, Ran, Xiao, and Wu}]{Hong2023MetaGPTMP}
Sirui Hong, Xiawu Zheng, Jonathan~P. Chen, Yuheng Cheng, Ceyao Zhang, Zili Wang, Steven Ka~Shing Yau, Zi~Hen Lin, Liyang Zhou, Chenyu Ran, Lingfeng Xiao, and Chenglin Wu. 2023.
\newblock \href {https://api.semanticscholar.org/CorpusID:260351380} {Metagpt: Meta programming for multi-agent collaborative framework}.
\newblock \emph{ArXiv}, abs/2308.00352.

\bibitem[{Huang et~al.(2023)Huang, Yu, Ma, Zhong, Feng, Wang, Chen, Peng, Feng, Qin, and Liu}]{Huang2023ASO}
Lei Huang, Weijiang Yu, Weitao Ma, Weihong Zhong, Zhangyin Feng, Haotian Wang, Qianglong Chen, Weihua Peng, Xiaocheng Feng, Bing Qin, and Ting Liu. 2023.
\newblock \href {https://api.semanticscholar.org/CorpusID:265067168} {A survey on hallucination in large language models: Principles, taxonomy, challenges, and open questions}.
\newblock \emph{ACM Transactions on Information Systems}, 43:1 -- 55.

\bibitem[{Islam et~al.(2023)Islam, Kannappan, Kiela, Qian, Scherrer, and Vidgen}]{Islam2023FinanceBenchAN}
Pranab Islam, Anand Kannappan, Douwe Kiela, Rebecca Qian, Nino Scherrer, and Bertie Vidgen. 2023.
\newblock \href {https://api.semanticscholar.org/CorpusID:265294665} {Financebench: A new benchmark for financial question answering}.
\newblock \emph{ArXiv}, abs/2311.11944.

\bibitem[{Li et~al.(2023)Li, Hammoud, Itani, Khizbullin, and Ghanem}]{Li2023CAMELCA}
G.~Li, Hasan Hammoud, Hani Itani, Dmitrii Khizbullin, and Bernard Ghanem. 2023.
\newblock \href {https://api.semanticscholar.org/CorpusID:257900712} {Camel: Communicative agents for "mind" exploration of large scale language model society}.
\newblock \emph{ArXiv}, abs/2303.17760.

\bibitem[{Liao and Vaughan(2023)}]{Liao2023AITI}
Qingzi~Vera Liao and Jennifer~Wortman Vaughan. 2023.
\newblock \href {https://api.semanticscholar.org/CorpusID:259075521} {Ai transparency in the age of llms: A human-centered research roadmap}.
\newblock \emph{ArXiv}, abs/2306.01941.

\bibitem[{Liu et~al.(2024)Liu, Xia, He, and Wang}]{Liu2024TrustworthinessAS}
Zhendong Liu, Changhong Xia, Wei He, and Chong-Jun Wang. 2024.
\newblock \href {https://api.semanticscholar.org/CorpusID:269804730} {Trustworthiness and self-awareness in large language models: An exploration through the think-solve-verify framework}.
\newblock In \emph{International Conference on Language Resources and Evaluation}.

\bibitem[{Luong et~al.(2024)Luong, Dinh, Raghavan, Nguyen, Nguyen, Le, Vo, Maegaito, Nguyen, Nguyen, Ha, and Nguyen}]{Luong2024DANADN}
Vinh Luong, Sang Dinh, Shruti Raghavan, Thanh Nguyen, Zooey Nguyen, Quynh Le, Hung Vo, Kentaro Maegaito, Loc Nguyen, Thao Nguyen, Anh~Hai Ha, and Christopher Nguyen. 2024.
\newblock \href {https://api.semanticscholar.org/CorpusID:273162306} {Dana: Domain-aware neurosymbolic agents for consistency and accuracy}.
\newblock \emph{ArXiv}, abs/2410.02823.

\bibitem[{Nguyen et~al.(2024{\natexlab{a}})Nguyen, Nguyen, Suzuki, Oku, Phan, Dinh, Nguyen, Ha, Raghavan, Vo, Nguyen, Nguyen, and Hirayama}]{Nguyen2024SemiKongCT}
Christopher Nguyen, Thanh Nguyen, Atsushi Suzuki, Daisuke Oku, Hong~An Phan, Sang Dinh, Zooey Nguyen, Anh~Hai Ha, Shruti Raghavan, Huy Vo, Thang Nguyen, Lan Nguyen, and Yoshikuni Hirayama. 2024{\natexlab{a}}.
\newblock \href {https://api.semanticscholar.org/CorpusID:274165766} {Semikong: Curating, training, and evaluating a semiconductor industry-specific large language model}.
\newblock \emph{ArXiv}, abs/2411.13802.

\bibitem[{Nguyen et~al.(2024{\natexlab{b}})Nguyen, Bi, Vosoughi, Tian, Fazli, and Xu}]{Nguyen2024OSCaROS}
Nguyen Nguyen, Jing Bi, Ali Vosoughi, Yapeng Tian, Pooyan Fazli, and Chenliang Xu. 2024{\natexlab{b}}.
\newblock \href {https://api.semanticscholar.org/CorpusID:268032097} {Oscar: Object state captioning and state change representation}.
\newblock In \emph{NAACL-HLT}.

\bibitem[{Nguyen et~al.(2024{\natexlab{c}})Nguyen, Tian, and Xu}]{Nguyen2024EfficientlyLL}
Nguyen Nguyen, Yapeng Tian, and Chenliang Xu. 2024{\natexlab{c}}.
\newblock \href {https://api.semanticscholar.org/CorpusID:268031847} {Efficiently leveraging linguistic priors for scene text spotting}.
\newblock \emph{ArXiv}, abs/2402.17134.

\bibitem[{Nguyen et~al.(2021)Nguyen, Nguyen, Tran, Tran, Ngo, Nguyen, and Hoai}]{Nguyen2021DictionaryguidedST}
Nguyen~Le Nguyen, Thua Nguyen, Vinh Tran, Minh-Triet Tran, Thanh~Duc Ngo, Thien~Huu Nguyen, and Minh Hoai. 2021.
\newblock \href {https://api.semanticscholar.org/CorpusID:235341891} {Dictionary-guided scene text recognition}.
\newblock \emph{2021 IEEE/CVF Conference on Computer Vision and Pattern Recognition (CVPR)}, pages 7379--7388.

\bibitem[{Park et~al.(2023)Park, O’Brien, Cai, Morris, Liang, and Bernstein}]{Park2023GenerativeAI}
Joon~Sung Park, Joseph~C. O’Brien, Carrie~J. Cai, Meredith~Ringel Morris, Percy Liang, and Michael~S. Bernstein. 2023.
\newblock \href {https://api.semanticscholar.org/CorpusID:258040990} {Generative agents: Interactive simulacra of human behavior}.
\newblock \emph{Proceedings of the 36th Annual ACM Symposium on User Interface Software and Technology}.

\bibitem[{Schick et~al.(2023)Schick, Dwivedi-Yu, Dess{\`i}, Raileanu, Lomeli, Zettlemoyer, Cancedda, and Scialom}]{Schick2023ToolformerLM}
Timo Schick, Jane Dwivedi-Yu, Roberto Dess{\`i}, Roberta Raileanu, Maria Lomeli, Luke Zettlemoyer, Nicola Cancedda, and Thomas Scialom. 2023.
\newblock \href {https://api.semanticscholar.org/CorpusID:256697342} {Toolformer: Language models can teach themselves to use tools}.
\newblock \emph{ArXiv}, abs/2302.04761.

\bibitem[{Senoner et~al.(2024)Senoner, Schallmoser, Kratzwald, Feuerriegel, and Netland}]{Senoner2024ExplainableAI}
Julian Senoner, Simon Schallmoser, Bernhard Kratzwald, Stefan Feuerriegel, and Torbj{\o}rn Netland. 2024.
\newblock \href {https://api.semanticscholar.org/CorpusID:270391431} {Explainable ai improves task performance in human–ai collaboration}.
\newblock \emph{Scientific Reports}, 14.

\bibitem[{Shinn et~al.(2023)Shinn, Cassano, Labash, Gopinath, Narasimhan, and Yao}]{Shinn2023ReflexionLA}
Noah Shinn, Federico Cassano, Beck Labash, Ashwin Gopinath, Karthik Narasimhan, and Shunyu Yao. 2023.
\newblock \href {https://api.semanticscholar.org/CorpusID:258833055} {Reflexion: language agents with verbal reinforcement learning}.
\newblock In \emph{Neural Information Processing Systems}.

\bibitem[{Wang et~al.(2023)Wang, Xie, Jiang, Mandlekar, Xiao, Zhu, Fan, and Anandkumar}]{Wang2023VoyagerAO}
Guanzhi Wang, Yuqi Xie, Yunfan Jiang, Ajay Mandlekar, Chaowei Xiao, Yuke Zhu, Linxi~(Jim) Fan, and Anima Anandkumar. 2023.
\newblock \href {https://api.semanticscholar.org/CorpusID:258887849} {Voyager: An open-ended embodied agent with large language models}.
\newblock \emph{Trans. Mach. Learn. Res.}, 2024.

\bibitem[{Wu et~al.(2023)Wu, Bansal, Zhang, Wu, Li, Zhu, Jiang, Zhang, Zhang, Liu, Awadallah, White, Burger, and Wang}]{Wu2023AutoGenEN}
Qingyun Wu, Gagan Bansal, Jieyu Zhang, Yiran Wu, Beibin Li, Erkang Zhu, Li~Jiang, Xiaoyun Zhang, Shaokun Zhang, Jiale Liu, Ahmed~Hassan Awadallah, Ryen~W. White, Doug Burger, and Chi Wang. 2023.
\newblock \href {https://api.semanticscholar.org/CorpusID:263611068} {Autogen: Enabling next-gen llm applications via multi-agent conversation}.

\bibitem[{Yang et~al.(2023)Yang, Yue, and He}]{Yang2023AutoGPTFO}
Hui Yang, Sifu Yue, and Yunzhong He. 2023.
\newblock \href {https://api.semanticscholar.org/CorpusID:259075577} {Auto-gpt for online decision making: Benchmarks and additional opinions}.
\newblock \emph{ArXiv}, abs/2306.02224.

\bibitem[{Yao et~al.(2022)Yao, Zhao, Yu, Du, Shafran, Narasimhan, and Cao}]{Yao2022ReActSR}
Shunyu Yao, Jeffrey Zhao, Dian Yu, Nan Du, Izhak Shafran, Karthik Narasimhan, and Yuan Cao. 2022.
\newblock \href {https://api.semanticscholar.org/CorpusID:252762395} {React: Synergizing reasoning and acting in language models}.
\newblock \emph{ArXiv}, abs/2210.03629.

\bibitem[{Zhang et~al.(2024)Zhang, Sun, Chen, Pfister, Zhang, and Arik}]{Zhang2024ChainOA}
Yusen Zhang, Ruoxi Sun, Yanfei Chen, Tomas Pfister, Rui Zhang, and Sercan~{\"O}. Arik. 2024.
\newblock \href {https://api.semanticscholar.org/CorpusID:270258551} {Chain of agents: Large language models collaborating on long-context tasks}.
\newblock \emph{ArXiv}, abs/2406.02818.

\end{thebibliography}

\end{document}